\RequirePackage{fix-cm}
\documentclass[twocolumn]{svjour3}          % twocolumn
\smartqed  % flush right qed marks, e.g. at end of proof
\usepackage{graphicx}
\usepackage{hyperref}
\usepackage[numbers]{natbib}
\journalname{KI - K{\"{u}}nstliche Intelligenz}
\usepackage[utf8]{inputenc}
\usepackage{wrapfig}

\newcommand{\quotesectiontitle}[1]{\begin{center}
\rule{\columnwidth}{0.4pt}
\bf \small \itshape
``#1''
\rule{\columnwidth}{0.4pt}
\end{center}
}

\newcommand{\KI}[1]{
\bigskip
\textit{ME: #1}
\bigskip}

\begin{document}
{\sloppy

\title{Intelligent behavior depends on the ecological niche \\
{\large Scaling up AI to human-like intelligence in socio-cultural environments} 
%{\large Interview with Dr.~Pierre-Yves Oudeyer} 
}

%\titlerunning{Interview Pierre-Yves Oudeyer}        % if too long for running head
\titlerunning{Intelligent behavior depends on the ecological niche}

\author{Manfred Eppe$^*$ \and Pierre-Yves Oudeyer}
% \author{Pierre-Yves Oudeyer}


\institute{$^*$Manfred Eppe \at
           Department of Informatics\\
           Universität Hamburg, Germany \\
           \email{eppe@informatik.uni-hamburg.de}}

\date{}

\maketitle

\begin{abstract}
This paper outlines a perspective on the future of AI, discussing directions for machines models of human-like intelligence. We explain how developmental and evolutionary theories of human cognition
should further inform artificial intelligence. We emphasize the
role of ecological niches in sculpting intelligent behavior, and in particular that
human intelligence was fundamentally shaped to adapt to a constantly changing
socio-cultural environment. We argue that a major limit of current work in AI is
that it is missing this perspective, both theoretically and experimentally.
Finally, we discuss the promising approach of developmental artificial intelligence,
modeling infant development through multi-scale interaction between intrinsically
motivated learning, embodiment and a fastly changing socio-cultural environment.
This paper takes the form of an interview of Pierre-Yves Oudeyer by Mandred Eppe,
organized within the context of a KI - K{\"{u}}nstliche Intelligenz special issue
in developmental robotics. 


\end{abstract}

%Abstract: In this interview, Pierre-Yves Oudeyer outlines his perspective on the future of AI, and discusses directions for machines models of human-like intelligence. He explains how neuro-cognitive and evolutionary theories of human cognition should further inform artificial intelligence. He emphasizes the role of ecological niches in sculpting intelligent behavior, and in particular that human intelligence was fundamentally shaped to adapt to a constantly changing socio-cultural environment. He argues that a major limit of current work in AI is that it is missing this perspective, both theoretically and experimentally.  Finally, he discusses the promising approach of developmental machine learning,
% modeling infant development through multi-scale interaction between intrinsically
% motivated learning, embodiment and a fastly changing socio-cultural environment. 

\section{Is it useful to talk about strong and weak AI?}

\KI{So what's the difference between strong and weak AI? What the broader public understands by AI is more the weak AI. So what is it that makes AI strong, and how is that related to your research?}

\textit{PYO:} That's an interesting question. First of all, I would like to say that I am not a big fan of the strong versus weak AI concept. 
I would say that for many people, when they speak about 'weak AI', they refer to techniques allowing to detect shallow relations between a number of quantities, and used to predict what will happen in the future without really understanding what's happening.
Whereas maybe some other people call 'strong AI' some kind of imaginary AI that is a bit like the concept of AGI -- Artificial General Intelligence. They intuitively define it as an intelligence that is going to be good at everything and be able to solve problems of unbounded complexity.

I don't like this distinction because, first of all, I don't think there is some kind of natural ordering of intelligences as is suggested by the 'strong versus weak' idea, which is some kind of linear scale in which some AI would be on top of another. It has the same problems as trying to order the intelligences of animals.
And second, I don't like it because I think that the AGI concept, behind the strong AI concept, is something that does not exist. Why so?

\quotesectiontitle{[W]hat counts is [...] to realize that each kind of intelligence is suited to its environment.}

I do believe, if you look at the world of living organisms in nature, that there is not one scale of intelligence: rather there is a diversity of intelligences. And each form of intelligence is adapted to the ecological niche in which it is living.
It has been evolving over quite long time to solve a particular family of problems that appear in their particular ecological niche. For example, ants have to solve problems which are very different from the ones solved by dogs which are very different from the ones solve by birds which are very different from the problems solved by humans.
And there is no natural ordering for this. So for example, the problems that ants are solving for being able to live in very extreme climatic and chemical environments is something that humans are very bad at. 
If you want to do an ordering from the point of view of ants, human intelligence is not so good. And it is the same for many other animals. 
For example, you have animals which need to have a very developed intelligence about figuring out the odor of things because they need to evolve in environments where vision is basically useless. 
And so they need to use senses related to smelling in a much more developed way than human beings. 
Again, from the point of view of these animals, the intelligence of other animals without those senses is basically very low. 
So what counts is actually not so much to classify the intelligences on a single scale but to realize that each kind of intelligence is suited to its environment.  

One key question behind this is: what is special about human intelligence? Rather than AGI, I prefer the concept of human-like intelligence (HLI), which is actually what some people understand intuitively when speaking about AGI. But human intelligence is not general at all. 
There are many, many problems at which we are very bad.
%\vfill\null 
% \columnbreak
% That's why school, for example, is difficult because at school, we learn many things which are difficult for many people, and which are actually very easy for some computers, by the way. 

% \quotesectiontitle{[T]o understand what is special about human intelligence, I think one needs to look at what is special about the ecological niche of the environment in which they are living.}

% \quotesectiontitle{One of the key features of the human environment is the social environment. And the social environment of humans is constantly changing.}

% \quotesectiontitle{This process of transmission causes high speed change of the social, cognitive, emotional environment to which all humans are exposed.
% And this continually evolving socio-cultural environment is essential in defining the problems humans need to solve.}
% \quotesectiontitle{[The] continually evolving socio-cultural environment [of humans] is essential in defining the problems humans need to solve.}

\section{Human-like intelligence}

Basically, to understand what is special about human intelligence, I think one needs to look at what is special about the ecological niche of the environment in which we are living. 
I think there are a few properties which are pretty important to understand the intelligence of humans. 
It's that their environment is characterized by others, by the social interactions with others, and that these social interactions continuously change through cultural evolution. 
Cultural evolution is basically the capability of human societies to not only learn novel concepts, novel ideas, invent new tools, but also  to share these inventions and transmit them to future generations. 
This process of transmission causes high speed change of the social, cognitive, emotional environment to which all humans are exposed.
And this continually evolving socio-cultural environment is essential in defining the problems humans need to solve\footnote{For discussions about the co-evolution of human cognition and socio-cultural structures, see \cite{tomasello2009cultural,donald1991origins,heyes2019precis,whiten2012human}}.
It's not like the ants, for example. The ants also have the rich social environment, but it is not constantly changing. There is no cultural evolution in ants. This is causing a special family of problems that humans need to solve, and in particular young humans who need to be able to learn continuously new rules for new games, new social games of interaction. This is a key dimension of the human environment, which I think is central to understand the human intelligence. And any machine aimed to be 'intelligent' in the human world would need to be able to deal with this rich fastly changing socio-cultural environment. 

\quotesectiontitle{Any machine aimed to be 'intelligent' in the human world would need to be able to deal with this rich fastly changing socio-cultural environment.}

Something that comes with cultural evolution is that humans are basically a species which is continuously making discoveries and inventions. 
This is thanks to the use of language, which is one of the main carrier of cultural evolution.
Language allows to transmit inventions to others, but it is also used as a tool to make discoveries\footnote{For discussions on various ways in which language is used as a cognitive tool, see \cite{Vygotskii1978,clark1998magic,gentner2016language}; For an artificial intelligence architecture modeling the use of language as a cognitive tool for driving creative discoveries in curiosity-driven exploration, see \cite{colas2020language}}).
That's very important because one of the key properties of human cognition is the ability to continuously invent new tools. 
Not only physical tools, like rakes or knives, but also cognitive tools such as natural languages, or formal languages, like mathematics. These cognitive tools have enabled the ratcheting effect of cultural evolution, which has enabled individuals to really expand the range of their discoveries.
I think the combination of cultural evolution and the ability to continuously invent new tools, in particular new cognitive tools, is something that is very, very peculiar to the human species. 

Human intelligence is basically optimized to be able to continuously adapt to its evolving cultural environment. And this cultural adaptation in turn leads to further change of the environment, forming a loop that is at the core of an open-ended growth of complexity. This probably leads to the need to be able to invent cognitive tools, to learn and negotiate new rules of social interaction. 
Understanding the mechanisms in this self-organizing complex system is one of the objectives of my approach to AI. 
I view AI as a tool that can enable us to understand the human species better. 
% To me, the first use of AI, from my own perspective, is really as a tool. In a way, I'm giving here some metastatements in the sense that I use AI as a formal language. 
As a cognitive tool enabling me to understand something about our human environment, and enabling me to grasp some of the special properties of humans: their capability to invent and use cognitive tools to understand better their (social) environment, and transmit them to others.

\section{The role of embodiment}

\quotesectiontitle{I don't think we can say very general statements about whether embodiment is needed for intelligence. I think it's rather that for some particular organisms in certain environments embodiment is absolutely central, and for other kinds of organism, in some other kinds of environments, maybe it is less central.}

\KI{Does this imply that an intelligent agent actually requires a body? For example, can there be intelligence in a chatbot? A chatbot doesn't really have a body, it has actions and perception and so on. But where are the limits and what kind of properties must an agent have to develop intelligent behavior? What's the minimum?}

Again, I think we need to put in perspective the question of whether something is needed for intelligence. 
I think it needs to be reframed in the context of the families of problems that a particular intelligence is supposed to solve in a particular family of environments. 
For example, human-like intelligence happens in the physical world and it's about controlling physical bodies in a physical environment with social beings in addition to physical objects.

This embodiment is a fundamental part of the problem solved by the intelligence. But it also is a fundamental part of the solution because the body itself has evolved through phylogenetic evolution in order to facilitate the capacity of the organism to adapt to the problems of its environment \cite{pfeifer2006body}. 
For example, if we look at language and we want to understand how humans understand language, we really quickly see that the properties of embodiment and how it is structurally aligned with the environment has a fundamental importance in the way to conceptualize meanings \footnote{See \cite{lakoff1999philosophy}, as well as \cite{cangelosi2010integration} for a research roadmap in developmental robotics studying the embodied language learning perspective.}. 
Many of the meanings that are expressed by language are defined in terms of the relationships between the body's repertoire of actions and the effects they can produce in the environment.
Meanings of sentences and meanings of words are relational entities that relate properties of the body and action repertoire with properties of the environment. 
Obviously, a system which has not a body with the fundamental properties of the specific body of humans will have very big difficulties to understand language in the same way as humans understand language.

Maybe in some other environments, there are entities which have to solve different families of problems in which embodiment is less essential. In that case, maybe embodiment is going to be less a problem and also less a solution because the world in which those entities are living is less relevant to physics. 
I don't think we can say very general statements about whether embodiment is needed for intelligence. I think it's rather that for some particular organisms in certain environments embodiment is absolutely central, and for other kinds of organism, in some other kinds of environments, maybe it is less central.

\KI{You say that social interaction like language and speech is, at least for human-level intelligence [...], a necessary property. Would you say that some artificial agent could be intelligent without any social interaction?}

Again, I would reformulate the other way around. I think that one of the key problems that human intelligence is solving is dealing with social interaction. 
% Social interaction, there are many animals who have it, but there is something very specific in this kind of social interaction that human have is cultural evolution, which is that the social norms continuously evolve with time. The preferences of others, the habits of others, the behavior of others continuously evolve with time.
% We as humans need to continuously adapt to those evolution of other individuals and evolution of the cultural environment in which we live. To me, the properties of being able to live in that world is one of the key problems that is solved by human intelligence.
Of course, you can imagine some other worlds. Actually, there are animals on this planet that have a rather poor social life and still they are able to do amazing things. It doesn't mean that they are not intelligent. From the point of view of their environment, they are extremely intelligent, they are extremely adaptive, it's just that they don't need to solve this problem of social interaction.
As I was saying initially, one of the specific problems that is implied by being able to live in a culturally evolving system is the ability to adapt continuously to changes in the environment and to invent new things. 

% \newpage

\section{Towards artificial learning systems with human-like intelligence}

\quotesectiontitle{I think what is missing in the current picture is an ecosystem [...] for evaluating machine learning systems [...] with respect to [...] targeting human-like intelligence.}

It's interesting that right now a big focus of the AI community is on machine learning, studying how artificial agents might be able to learn novel things. 
The frontier of this field is, in a way, life-long learning, to continuously learn new things in an extended period of time.
I think what is missing in the current picture is an ecosystem that is relevant for training and then evaluating those machine learning systems and assessing their relevance, for example, with respect to targeting human-like intelligence. 
If you look, for example, at most of the benchmarks that currently exist in the field of deep reinforcement learning, which is about how to make sequences of decisions to solve problems:
How many of them involve solving complex behavioral problems related to \textit{social interaction}? Very little! 
It's only very recently that people have began to become interested in language grounding in this community of machine learning. Things are developing, but it is still very primitive as compared to the real social problems that are solved by humans.

\KI{It would be a very valuable contribution, for instance, for the development of robotics, to develop new benchmark environments which feature this open-ended learning, right?}

Yes. Developmental robotics, long before machine learning, has for more than two decades being focusing on those problems of how machines can adapt and learn interaction with social peers \cite{cangelosi2015developmental,kaplan2008computational,tani2016exploring}.
People have studied for those two decades, for example, how children can learn basic social interaction skills such as joint attention \cite{breazeal2002robots,kaplan2006challenges,nagai2016mechanism}. 
Also, the problem of language grounding has been studied already 20 or 30 years ago, even before developmental robotics started as a field \cite{siskind1994grounding, steels2001language,roy2002learning}.

What's interesting to see is that, initially, those concepts were rather studied from the point of view of really modeling directly human intelligence. 
I think this was really the right way to go. Generally, machine learning has ignored all these conceptual thinking, which is mixing computational modeling, developmental psychology, developmental cognitive neuroscience. They have ignored this for a very long time and they are slowly rediscovering many of those old concepts, which are very, very fundamental.
Hopefully now, by leveraging and rediscovering these concepts, there is a great opportunity for AI: very modern and efficient machine learning techniques that exist, but which so far have not been developed and assessed in the right environmental context, could now be adapted and used in the right ecosystem of problems that are relevant for humans. This emerging convergence between developmental robotics and machine learning, which I like to call "developmental machine learning", or "developmental AI", is really exciting!
\vfill\null
%\columnbreak
\section{Interdisciplinary synergies between AI and cognitive sciences}

\quotesectiontitle{This emerging convergence between developmental robotics and machine learning, which I like to call "developmental machine learning", or "developmental AI", is really exciting!}

% \quotesectiontitle{What we actually discovered is that there are certain curiosity mechanisms, [...]
% when you use them in an embodied system and you let this system run for a long period of time, it self-organizes developmental trajectories with phases of increasing complexity that have special properties not preprogrammed in the system.}

\KI{Obviously, neuroscience, psychology, and cognitive science has learned a lot from AI and vice versa. What do you think are the biggest chances here? What does AI have that neurosciences, psychology, and cognitive sciences can learn most from. Or the other way around. Can you give an example?}

There are many things \footnote{For a review, see \cite{oudeyer2010impact}}. I will use my own work as an example. Psychologists have been discussing the idea that the brain might be motivated to explore novel situations or changing situations for many, many decades. Already in the '40s and the '50s, there were psychologists like Daniel Berlyne who proposed to conceptualize curiosity \cite{berlyne1960conflict}. They proposed that it might be a very fundamental mechanism to understand how children grow up and develop intelligence.
For many decades, this concept remained expressed in a verbal manner, without being very well formalized. 
As a consequence, it was little considered in psychology, neuroscience and AI, even though Berlyne, Piaget and others said that it is probably very important for the long-term cognitive development of children. 
Also, it had not clearly been identified what could be its potential role in learning and development. 

With a few labs in the world, when we began at the beginning of the 21st century to study and to model the concept of curiosity, what we were really interested in was to understand the underlying mechanisms and how they could drive aspects of the development of cognition and sensorimotor skills \footnote{See \cite{baldassarre2013intrinsically,oudeyer2007intrinsic} for early modeling works on this topic}. 
The first thing we did was to formalize the space of potential mechanisms \cite{oudeyer2009intrinsic} and then to develop a series of experiments to try to understand the consequences of using these kinds of mechanisms. 
We discovered that one form of curiosity mechanisms, which we named the 'learning progress hypothesis' had very interesting properties\cite{kaplan2007search}: these mechanisms basically lead the brain to explore learning situations (e.g. actions, goals or people) which enable to improve predictions or control about the world.
What we actually discovered is that these curiosity mechanisms, when you use them in an embodied system and you let this system run for a long period of time, it self-organizes developmental trajectories with phases of increasing complexity that have special properties not preprogrammed in the system.
At the beginning, there was no blueprint of these schedules. 
These schedules are really something emergent and we were able to discover that they share many structural properties with infant development of sensorimotor and cognitive skills. 
For example, for the case of vocal development, we were able to show that if you put this system into an artificial vocal tract, it will automatically generate a curriculum of learning \cite{moulin2014self}. 
The phases are very, very similar to the phases of vocal development we observed in infants. We later conducted other experiments showing how such curiosity could lead to the discovery of complex forms of tool use, also with similarities with human developmental trajectories \cite{forestier:hal-01583301}.

Basically, this kind of work enabled not only to generate precise hypotheses about the mechanisms of curiosity. It also enabled to show, to propose the hypothesis that the moment-to-moment mechanism of curiosity, when unrolled on the long-term in an organism, enables self-organization of developmental structure, with the emergence of major behavioral and cognitive transitions \cite{oudeyer2016evolution}. 
This could actually explain some of the developmental structures that we know are existing in humans. Before, there was no theory that could explain how they arise.
% Basically, we knew a number of things happen in a certain order with certain regularities and also with certain diversity among individuals. We did not know why. 
Now, after we've done this computational modeling work, we have those hypotheses. We are not sure it's actually working like this in humans, but now, we have those hypothesis and we can design new experiments to test those predictions. And this is exactly what us and other labs in the world are currently doing  \footnote{See 
\cite{gottlieb2018towards,kidd2015psychology} for reviews of recent developments in psychology and neuroscience}. This is an example of how AI can be used as a tool to understand better how humans are working.

\quotesectiontitle{What we actually discovered is that these curiosity mechanisms, when you use them in an embodied system and you let this system run for a long period of time, it self-organizes developmental trajectories with phases of increasing complexity that have special properties not preprogrammed in the system.}

I think also AI has been taking already a lot of insights from cognitive sciences for many decades and many others are still to be used. For example, a lot in the past has been done about the structure of memories. 
Cognitive science, psychology and neuroscience have been knowing for a very long time that in the human brain, there are different forms of memories. These have different functionalities and different structures. These ideas have inspired several generations of AI researchers to put different kinds of memories in their systems.
For example, when we look right now at deep reinforcement learning systems \footnote{See \cite{hassabis2017neuroscience} for a review on how recent work in deep learning takes inspiration from neuroscience.}, they need some forms of episodic memories. For example, in the form of replay buffers with other forms of more parametric memories or more declarative memories. That's a bit like what's happening in the human brain. 

\quotesectiontitle{You have many people right now who are focusing on enabling an AI system to master a very complicated adult-like language, but very little people are actually trying to see how one can enable complex social reasoning capacities in artificial systems, which are there well before language in infants.}

I think that this is something that was initially inspired a lot from cognitive sciences. 
There are other things which cognitive sciences, and psychology, and neuroscience have known for a long time and which are still very little exploited in AI and machine learning.
For example, if we go back to the question of cultural evolution and social interaction, we know that there are important complex skills that are developed by very young children. 
For example, the ability to build mental models of others, to have a motivation to help others, and to be altruistic in social interaction. 
Michael Tomasello \cite{tomasello2009cultural}, among others, has shown that very, very early on, infants who have very limited understanding of language, they can already understand in a very advanced manner social situations in which they guess the goals of others. They can anticipate what others are going to do next, and they can spontaneously propose to help them. These are incredible feats of the human mind.
It is amazing that humans are doing this even before they master advanced language. 

You have many people right now who are focusing on enabling an AI system to master a very complicated adult-like language, but very little people are actually trying to see how one can enable complex social reasoning capacities in artificial systems, which are there well before language in infants.
By the way, maybe very important for language grounding, it is not only key to understand how meanings are grounded in relations between the body and physical objects, but also in relations between oneself and others in a social context with joint goals. Language is a tool that enables humans to achieve joint goals. 
To understand language and its use, it seems it's quite important that agents understand the concept of joint goals and coordination with others.

\section{Future directions and conclusion}

\quotesectiontitle{I have the feeling that there is something to discover about the importance of cognitive bottlenecks.}

\KI{Thanks a lot, and I have one concluding question. If you were a young person just starting to study, having some knowledge about computer science and maybe psychology, what would you start right now? If you knew nothing of what you know now, or maybe only a little bit, what would be the topic that you would tackle?}

% Interviewer: Thanks a lot. I think we are almost stopping now. I have one concluding question. Otherwise, it all gets too long. If you have a final comment, if you are now a very young person just starting to study, have some knowledge on computer science and maybe psychology, what would you start now? If you knew nothing that you know now or maybe only a little bit, what would be the topic that you would start now?

That's a tough question because there are so many interesting things. There are dozens of PhD topics I would like to do. The difficult thing would be to choose one. Rather than giving you the one I think is best, maybe I give you an example of what I have been thinking about with my team, and which I think is very interesting research.
Some of the properties of human cognition are a bit mysterious and strange. For example, if you look at properties of working memory, there is a bottleneck in human working memory. We cannot hold and reason with more than four to six, seven objects or concepts at the same time. 
Doing that reasoning with more objects or concepts or entities is very, very difficult for our brain. 
This is very strange because our brain is so powerful. We have long-term memory that is able to retain literally a huge amount of things, way more than a machine is able to do today. We are also able to do some amazing long-term reflection and inferences, but yet at the same time, we cannot hold in our head much more than five, six objects.
Why is this so? Especially, if you look at animals, for example, if you look at certain non-human primates. 
Many of them have a much better working memory than we have. Maybe you know this experiment with monkeys, where you put them in front of a screen and you show numbers at different scrambled places of the screen from 1 to 10. You show this image for a few seconds and then it's blank. Then you only show the empty squares on the screen and you need to type in the numbers they contained initially. This is completely impossible to do for humans, but that's very easy for many monkeys. They can remember, for a short amount of time, a larger number of entities than what we can. Yet humans have language, humans have cultural evolution, humans have invented mathematics, humans have literature, and poetry.
I have an intuition about this, and I'm not the only one. There is a number of people who have also outlined this intuition, for example, the famous primatologist Tetsuro Matsuzawa. 
He has proposed the cognitive tradeoff hypothesis \cite{matsuzawa2009symbolic}, which states that this bottleneck in the working memory of humans might be some kind of constraint that has pushed or caused in some way the emergence of a number of cognitive tools like language, compositionality, and all the tools in our cognitive toolkit that enabled the combinatorial discoveries that we make. 

I have the feeling that there is something to discover about the importance of cognitive bottlenecks. We have huge computational resources, and many people in AI focus on trying to exploit it. But on the contrary, if we have some constraints on the computational resources, then maybe it can have some advantage in terms of developing an architecture which is going to be very powerful in the environment of humans. 
What I'm saying here is still not very precise, but I feel there is something very interesting to study here. 
That would be an example of a direction that could be very fruitful to study further.

\KI{OK, I conclude then that more PhD students should start working in this field and maybe be not so overwhelmed by AI and machine learning hype, but also look more into the deeper cognitive mechanisms that make humans intelligent.}

Yes, exactly.

\KI{Great! Thank you very much for the interview!}

\section{Short Bios}

\noindent Pierre-Yves Oudeyer is a research director at Inria and head of the Flowers lab at Inria, Univ. Bordeaux and Ensta-ParisTech since 2008. The Flowers team hosts around 25 researchers at the crossroads of AI, machine learning, developmental robotics and cognitive sciences. From 1999 until 2007, he was a permanent researcher in Sony Computer Science Laboratory.
\begin{wrapfigure}[12]{l}{0.46\columnwidth}
\vspace{-15pt}
\includegraphics[width=0.49\columnwidth]{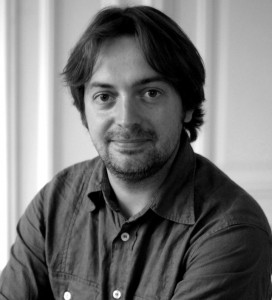}
\end{wrapfigure} 
He studies developmental autonomous learning and the self-organization of behavioral and cognitive structures in humans and machines. In particular, he studies exploration in large open-ended spaces, with a focus on autonomous goal setting, intrinsically motivated learning, and how this can automate curriculum learning. With his team, he pioneered curiosity-driven learning algorithms working in real world robots (used in Sony Aibo and Qrio humanoid robots), and showed how the same algorithms can be used to personalize sequences of learning activities in educational technologies deployed at large in schools. He developed theoretical frameworks to understand better human curiosity and its role in cognitive development, and contributed to build an international interdisciplinary research community on human curiosity. He also studied how machines and humans can invent, learn and evolve speech communication systems. 

PY Oudeyer has received multiple international awards, such as the 2018 Prize Inria from National Academy of Science (France), the 2016 Lifetime Achievement Award from the Evolutionary Linguistics association, and an ERC Starting Grant (2009). He is associate editor of IEEE CIS Newsletter on Cognitive and Developmental Systems, the IEEE Transactions on Cognitive and Developmental Systems and Frontiers in Neurorobotics. He is also actively contributing to disseminate science towards the general public, by writing popular science articles and participating in radio and TV programs as well as science exhibitions.

\noindent Dr. Manfred Eppe is a principal investigator and senior researcher at the Knowledge Technology Group at  Universität Hamburg, Germany, since 2017. From 2015 to 2017 he was a postdoctoral research fellow at the International Computer Science Institute at the University of California at Berkeley, USA. He was a postdoctoral researcher at the Artificial Intelligence Research Institute, Barcelona, Spain, from 2014 to 2015 and obtained his Ph.D. from the University of Bremen, Germany, in 2014. 
\begin{wrapfigure}[12]{l}{0.46\columnwidth}
\vspace{-15pt}
\includegraphics[width=0.49\columnwidth]{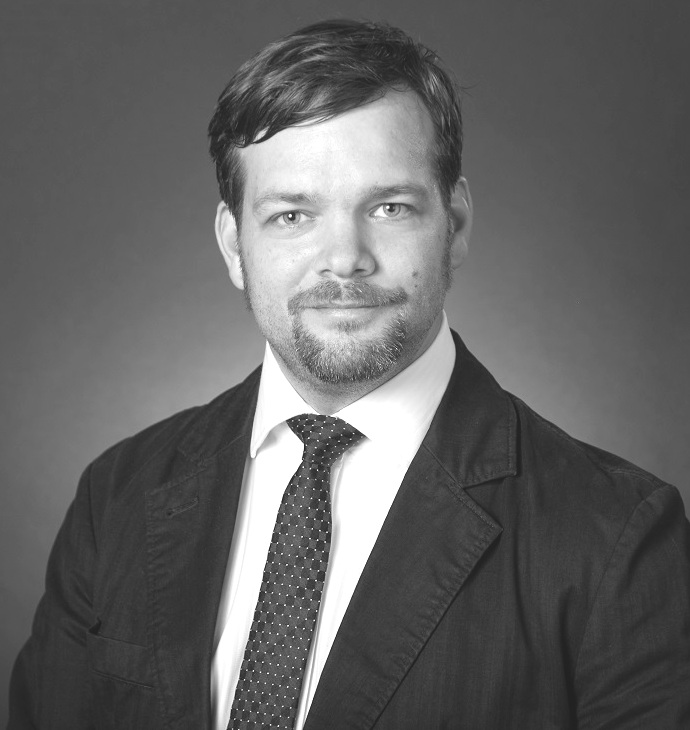}
\end{wrapfigure} 
Manfred Eppe is a regular reviewer and author of papers for conferences in robotics, artificial intelligence, machine learning and cognitive science, including IROS, ICDL-Epirob, ICANN, IJCAI, Transactions on Cognitive and Developmental Systems, IJCNN, Frontiers in Robotics and AI, and Artificial Intelligence. He has led or is leading several research projects funded by the Volkswagen Stiftung, the German Academic Exchange Service (DAAD) and the German Research Foundation (DFG). 
Manfred Eppe studies the links between Cognitive Sciences and Artificial Intelligence. 
His main interest is in representation learning and concept grounding: How can embodied (robotic) agents learn to represent their environment as a composition of abstract concepts, and to perform decision-making based on such abstract representations?
He is particularly active in the fields of reinforcement learning, self-supervised learning and natural language processing, combining symbolic and neural network-based methods of artificial intelligence to model cognitive processes. 
Manfred Eppe has interviewed Pierre-Yves Oudeyer about his perspective on the future of more human-like artificial intelligence.

%KI spoke with him about the future of Artificial Intelligence, and especially how neurocognitive theories about decision-making, action and perception can help to improve the current machine learning-based approaches that sometimes are referred to as `weak AI'. 

} % end sloppy

\bibliographystyle{plainnat}
\bibliography{references}

\end{document}